\documentclass[11pt]{article}

\usepackage[final]{acl}
\usepackage{times}
\usepackage{latexsym}
\usepackage[T1]{fontenc}
\usepackage[utf8]{inputenc}
\usepackage{microtype}
\usepackage{inconsolata}
\usepackage{graphicx}
\usepackage{amsmath}
\usepackage{bm}
\usepackage{stfloats}
\usepackage{booktabs}
\usepackage[para]{threeparttable}
\usepackage{multirow}
\usepackage{float}
\usepackage{url}

\DeclareUnicodeCharacter{2212}{-}

\newif\ifanonymous
\anonymousfalse

\title{MGhana-ST: A Low-Resource Speech Translation Dataset for Ghanaian Languages and an Analysis of Multilingual Training Trade-offs}

\ifanonymous
  \author{Anonymous submission}
\else
  \author{
    Frank Lawrence Nii Adoquaye Acquaye \\
    Ashesi University, Berekuso, Ghana \\
    AdwumaTech AI, Accra, Ghana \\
    \texttt{facquaye@ashesi.edu.gh}
    \And
    Eric George Parakal \\
    HSE University, Moscow, Russia \\
    \texttt{ericparakal@gmail.com}
    \AND
    Jesse Johnson \\
    AdwumaTech AI, Accra, Ghana \\
    \texttt{jesse.johnson@adwumatech.ai}
    \And
    Kishankumar Bhimani \\
    GIFT International Fintech Institute, \\
    Gandhinagar, India \\
    \texttt{info.bhimani@gmail.com}
    \AND
    Jochebed Afua Basil \\
    Ashesi University, Berekuso, Ghana \\
    \texttt{jochebed.basil@ashesi.edu.gh}
  }
\fi

\begin{document}
\maketitle

\begin{abstract}
We present MGhana-ST, a speech translation dataset for four low-resource Ghanaian language varieties: Ga, Twi (Akuapem and Asante), Ewe, and Fante. MGhana-ST is an ongoing annotation effort; the experiments reported here use a fixed subset of approximately 16.1 hours of paired speech and English translation data. The audio is curated from two existing Ghanaian speech resources. The English translations in MGhana-ST are produced directly from the audio by 37 native-speaker annotators, rather than derived from source-language transcriptions as in the upstream resources, and are accompanied by verbal and non-verbal event annotations.

Using Whisper-small as a backbone, we examine monolingual and multilingual training under severe data scarcity, reporting all baseline results as means over three training seeds. We find that flat multilingual training benefits no variety in this regime. Ga and Twi are unchanged within seed variance ($+0.51$ and $+0.06$ BLEU against monolingual standard deviations of $1.63$ and $2.20$), while Ewe declines by $6.99$ BLEU and Fante by $5.11$. The two varieties that degrade are Ewe, which is both linguistically distinct and drawn from a different source corpus, and Fante, the least-resourced variety. We further compare empirical cross-lingual transfer with typology-based similarity and find that, in this four-language setting, transfer BLEU identifies which language pairs interact more closely than URIEL similarity does, although neither measure predicts which varieties benefit from joint training.

We also report a methodological finding of independent interest. An earlier single-run version of this analysis found positive transfer for three of four varieties; that result did not survive replication across seeds. For Ga and Twi, monolingual baselines trained on roughly 1.6 to 6.2 hours of audio have seed standard deviations of $1.63$ and $2.20$ BLEU, roughly five and thirty times those of the corresponding multilingual models ($0.35$ and $0.07$). When the monolingual condition is the noisier one, a single-run comparison can show apparent transfer of this size from seed variation alone. We release MGhana-ST to support future research on African language speech technologies and low-resource multilingual speech translation.
\end{abstract}

\section{Introduction}
Multilingual speech models have significantly expanded the scope of automatic speech recognition and speech translation. Whisper \citep{radford2023robust}, trained on 680{,}000 hours of audio across 99 languages, achieves strong performance on a range of speech benchmarks, while the Massively Multilingual Speech project \citep{pratap2024scaling} extends multilingual speech modelling to more than 1{,}100 languages. However, scaling multilingual training also introduces a central tension: parameter sharing across languages can improve performance through positive transfer, but it can also degrade performance through negative transfer when jointly trained languages are insufficiently compatible \citep{wang2020negative,arivazhagan2019massively}. This trade-off is especially consequential in low-resource settings, where the amount of supervised data is too limited to absorb poorly matched multilingual signals.

Speech translation (ST) enables direct translation from speech in a source language to text in a target language without requiring an intermediate automatic speech recognition (ASR) step \citep{berard2016listen,duong2016attentional}. In our setting, reliable ASR support for Ga, Twi, Ewe, and Fante remains limited, making end-to-end speech translation a practical modelling choice. The subset used in our experiments comprises about 16.1 hours of labelled speech-translation data across the four varieties, a regime in which model design and multilingual training strategy can materially affect final system quality.

The Ghanaian languages considered here provide a useful testbed for studying multilingual transfer under partial linguistic relatedness. Twi (in its Akuapem and Asante varieties) and Fante belong to the Akan branch and are more closely related to one another than either is to Ga or Ewe (see the language family column in Table~\ref{tab:full-data}). Relatedness within this set is uneven: some language pairs are more likely to benefit from multilingual transfer, while others may introduce cross-language interference. These properties make the setting well-suited to studying when multilingual speech translation helps, when it hurts, and whether task-specific transfer signals are more informative than typological similarity measures for designing multilingual training configurations.

Beyond introducing MGhana-ST as a new resource for low-resource Ghanaian speech translation, we use the dataset as a controlled benchmark for analysing multilingual training under severe data scarcity. We distinguish between the dataset itself, which is intended as a reusable community resource, and the empirical observations obtained from our experimental study. The latter should be interpreted as evidence from the modelling choices considered here rather than as intrinsic properties of the dataset.

Our experiments yield four main observations. First, flat multilingual training benefits no variety at this data scale: Ga and Twi are unchanged within seed variance, while Ewe and Fante both degrade substantially. Second, the varieties that degrade share identifiable characteristics: Ewe is the only Gbe variety and is also drawn from a different source corpus, while Fante is the least-resourced variety (1.62h). These observations are consistent with the possibility that joint training reallocates limited capacity toward better-represented varieties, although our experiments do not isolate this mechanism. Third, empirically measured cross-lingual transfer BLEU identifies which language pairs interact more closely than URIEL typological similarity does within our four-language setting, but it does not predict which varieties benefit under joint training. Fourth, and methodologically, monolingual baselines at this scale carry seed standard deviations of up to $2.20$ BLEU, comparable to the single-run gains of $+1.68$ to $+2.95$ BLEU that our own earlier analysis reported; single-run comparisons are therefore not adequate in this regime.

Our contributions comprise two distinct parts: a new dataset resource and an empirical study conducted on it. We keep these separate throughout, and readers interested only in the resource may read Section~\ref{sec:dataset} in isolation.

\paragraph{Dataset contribution.}
\begin{itemize}
    \item \textbf{Audio-grounded translation and event annotation:} All English translations in MGhana-ST were produced by 37 native-speaker annotators listening to the audio, together with verbal and non-verbal event tags.
    \item \textbf{Dataset Release:} We publicly release MGhana-ST, a multi-language Ghanaian speech translation resource covering Ga, Twi, Ewe, and Fante, together with per-language corpus statistics and an explicit manifest defining the experimental subset.
\end{itemize}

\paragraph{Empirical study.}
\begin{itemize}
    \item \textbf{Multilingual Training Analysis:} We compare monolingual and multilingual speech translation models for four low-resource Ghanaian language varieties across three training seeds.
    \item \textbf{Empirical Transfer Study:} We analyse cross-lingual transfer within our language set and compare transfer-based and typology-based relatedness measures.
    \item \textbf{Seed sensitivity at 16 hours:} We quantify training-seed variance for both monolingual and multilingual configurations and show that it is large enough to reverse the sign of reported transfer effects.
\end{itemize}

\section{Related Work}

Large-scale multilingual speech models such as Whisper \citep{radford2023robust}, MMS \citep{pratap2024scaling}, and mSLAM \citep{bapna2022mslam} establish broad multilingual capability across many languages and tasks. More recent work, including SpeechMatrix \citep{duquenne2023speechmatrix} and SeamlessM4T \citep{barrault2023seamlessm4t}, extends multilingual speech translation to a large number of languages within unified models. However, these systems primarily focus on scaling capabilities across many languages rather than investigating the specific dynamics of positive and negative transfer within small sets of partially related varieties at severe data scarcity.

Speech resources covering African languages have expanded in recent years, including massively multilingual benchmarks such as FLEURS \citep{conneau2023fleurs}, accented-speech recognition corpora such as AfriSpeech-200 \citep{olatunji2023afrispeech}, and speech synthesis corpora such as BibleTTS \citep{meyer2022bibletts}. For Ghanaian languages specifically, UGSpeechData \citep{wiafe2025ugspeechdata} and the Financial Inclusion Speech Dataset \citep{owusu2022financial} provide transcribed speech. Most of these resources were designed primarily for ASR or speech synthesis rather than end-to-end speech translation. MGhana-ST addresses this gap by providing an audio-grounded speech translation dataset for four Ghanaian language varieties and an analysis of multilingual training under severe data scarcity. Our positioning is complementary: while large-scale systems demonstrate broad capability, this work examines the fine-grained transfer dynamics that those aggregate evaluations do not reveal, using typology-based similarity (URIEL) and empirical transfer BLEU as alternative lenses on language relatedness.

\section{The MGhana-ST Dataset}
\label{sec:dataset}

MGhana-ST is a curated dataset of short speech clips in Ghanaian languages paired with English translations and annotations for non-verbal events such as laughter, applause, and background noise. It is available at
\ifanonymous
  \texttt{[URL withheld for review]}
\else
  \url{https://huggingface.co/datasets/adwumatech-ai/mghana-st}
\fi
\citep{acquaye2025mghana}.

\paragraph{Licensing.}
The annotation layer contributed by this work, comprising the English translations and the verbal and non-verbal event tags, is released under the Creative Commons Attribution 4.0 International licence (CC BY 4.0). The underlying audio remains under the terms set by its original distributors: the Financial Inclusion Speech Dataset is distributed under CC BY 4.0 \citep{owusu2022financial}, and Ewe audio from UGSpeechData under CC BY-NC-ND 4.0 \citep{wiafe2025ugspeechdata}, which restricts it to non-commercial use.

\subsection{Defining the Experimental Subset}
\label{sec:subset-definition}

MGhana-ST is an incremental project: annotation is ongoing, and further audio is being processed for inclusion. The \emph{experimental subset} is the fixed portion of the data used for every result reported in this paper, summarised in Table~\ref{tab:full-data}; it comprises approximately 16.1 hours across the four varieties. The \emph{public release} is the annotated data available at the repository above.

\paragraph{The subset is defined by manifest, not by revision alone.}
The annotation state underlying our experiments corresponds to commit \texttt{4996b709abd584464a6c\allowbreak fb7bd256504fd5eed7a0} of the dataset repository, dated 25 March 2026. We release a manifest at \texttt{splits/manifest.json} in the dataset repository, listing the \texttt{audio\_id}, \texttt{start\_ms}, and \texttt{end\_ms} of every segment assigned to train, validation, and test. This manifest is the authoritative description of what our experiments used.

\paragraph{Audio availability.}
Of the 90{,}583 annotated segments across the four varieties, 11{,}445 (12.6\%) reference audio files that were not present locally and therefore never entered preprocessing. Their distribution is highly uneven across varieties: 39.9\% of Ewe segments and 30.4\% of Fante segments, against 0.03\% for Twi and none for Ga (Table~\ref{tab:retention}, Appendix~\ref{sec:exclusion}). We have not determined whether these files are absent from the upstream distributions or absent only from our copy, nor whether their absence is systematic. The Ewe and Fante portions of the experimental subset should therefore be understood as samples of their annotated material whose selection we do not fully characterise.

\subsection{Sources}

MGhana-ST aggregates audio from two existing Ghanaian speech resources. Ewe samples were drawn from the transcribed portion of UGSpeechData \citep{wiafe2025ugspeechdata}, a multilingual Ghanaian speech dataset covering Akan, Ewe, and other languages. The remainder was drawn from the Financial Inclusion Speech Dataset \citep{owusu2022financial}, which covers Akuapem Twi, Asante Twi, Fante, and Ga and leans heavily on the financial technology domain.

Corpus provenance is not evenly distributed across language varieties: all Ewe data is drawn from UGSpeechData, whereas Ga, Twi, and Fante are drawn entirely from the Financial Inclusion Speech Dataset. Because the two corpora differ systematically in domain, recording channel, speaking style, and utterance duration, Ewe differs from the other three varieties along these dimensions as well as linguistically. We return to the implications of this in Section~\ref{sec:results} and the Limitations.

\paragraph{What is new in MGhana-ST.}
The source audio is reused from the two corpora above. The annotation layer is not. MGhana-ST provides translations produced by native speakers listening to the audio directly, rather than translations of written transcriptions as in the upstream resources. Annotators additionally tagged verbal and non-verbal events (laughter, applause, background noise), which have no counterpart upstream.

\subsection{Languages and Coverage}
\label{sec:coverage}

The experimental subset covers four Ghanaian language varieties: Twi, Ewe, Ga, and Fante. Table~\ref{tab:full-data} reports its train, validation, and test durations per variety. Twi and Fante belong to the Akan branch and are closely related, whereas Ga (Ga-Dangme) and Ewe (Gbe) are more distant from the Akan varieties and from each other.

\paragraph{Split methodology.}
Splits are constructed at the audio-file level: all segments originating from a single recording are assigned to exactly one of train, validation, or test, so no audio used for training appears in validation or test.

\paragraph{Corpus characteristics.}
Table~\ref{tab:corpus-stats} reports per-language utterance counts and durations. Utterances are short: mean duration is 1.33\,s for Fante, 1.68\,s for Ga, and 1.90\,s for Twi. Ewe, drawn from image-description recordings, averages 3.31\,s. English reference diversity differs sharply across varieties (Table~\ref{tab:reference-diversity}): 38.4\% of Fante test references repeat another reference, against only 1.5\% for Ewe. MGhana-ST should therefore be understood as a word- and phrase-level speech translation resource for three of its four varieties, and a mixed phrase- and sentence-level resource for Ewe.

\begin{table}[ht]
\centering
\small
\setlength{\tabcolsep}{4pt}
\begin{tabular}{llrrr}
\hline
\textbf{Language} & \textbf{Family} & \textbf{Train (h)} & \textbf{Val (h)} & \textbf{Test (h)} \\
\hline
Twi         & Akan      & 6.21 & 0.87 & 0.79 \\
Ga          & Ga-Dangme & 2.48 & 0.29 & 0.27 \\
Ewe         & Gbe       & 2.47 & 0.37 & 0.34 \\
Fante       & Akan      & 1.62 & 0.19 & 0.18 \\
\hline
\textbf{Total} &        & 12.78 & 1.72 & 1.58 \\
\hline
\end{tabular}
\caption{The MGhana-ST experimental subset: train, validation, and test audio hours per language. This is the fixed subset used for all experiments reported here.}
\label{tab:full-data}
\end{table}

\begin{table*}[t]
\centering
\small
\begin{tabular}{lrrrrrrr}
\hline
\textbf{Language} & \textbf{Family} & \textbf{Train} & \textbf{Val} & \textbf{Test} & \textbf{Total} & \textbf{Total} & \textbf{Mean dur.} \\
 & & \textbf{utts.} & \textbf{utts.} & \textbf{utts.} & \textbf{utts.} & \textbf{(h)} & \textbf{(s)} \\
\hline
Twi   & Akan      & 11{,}849 & 1{,}593 & 1{,}484 & 14{,}926 & 7.87 & 1.90 \\
Ga    & Ga-Dangme &  5{,}278 &   624 &   639 &  6{,}541 & 3.05 & 1.68 \\
Ewe   & Gbe       &  2{,}714 &   405 &   344 &  3{,}463 & 3.18 & 3.31 \\
Fante & Akan      &  4{,}328 &   522 &   524 &  5{,}374 & 1.99 & 1.33 \\
\hline
\end{tabular}
\caption{Per-language corpus characteristics of the experimental subset. Utterance counts include all annotations; where a clip carries more than one independent translation, all references are retained. Total hours may differ from the sum of the per-split hours in Table~\ref{tab:full-data} by 0.01\,h due to rounding.}
\label{tab:corpus-stats}
\end{table*}

\begin{table}[t]
\centering
\small
\begin{tabular}{lrrr}
\hline
\textbf{Language} & \textbf{Unique} & \textbf{Target} & \textbf{Multi-ref.} \\
 & \textbf{refs.} & \textbf{overlap} & \textbf{rate} \\
\hline
Twi   & 1{,}078 & 27.4\% & 2.7\% \\
Ga    &   465 & 27.2\% & 0.6\% \\
Ewe   &   339 &  1.5\% & 0.3\% \\
Fante &   323 & 38.4\% & 4.2\% \\
\hline
\end{tabular}
\caption{Reference diversity on the test split. Target overlap is the share of test utterances whose English reference is not unique ($1 - \text{unique refs.}/\text{test utts.}$, with test utterance counts from Table~\ref{tab:corpus-stats}). Ewe references are almost entirely unique, whereas over a third of Fante references are repeated.}
\label{tab:reference-diversity}
\end{table}

\subsection{Annotation and Quality Assurance}
\label{sec:annotation}

All English translations and non-verbal event tags were produced by 37 annotators, each a native speaker of the language they annotated. All annotators were employed and paid by AdwumaTech AI and gave informed consent for their translations and annotations to be released. Quality assurance operated at three levels: redundant annotation (a subset of clips received multiple independent translations; 0.3--4.2\% of test utterances per variety, Table~\ref{tab:reference-diversity}), guided native-speaker annotation (teams working from written guidelines covering event annotation and idiomatic expression handling), and two-round independent verification by more experienced annotators. As an independent check on annotation hygiene, we audited every bracketed event tag in the corpus. Of 56{,}117 translations containing an event tag, 36 (0.06\%) are malformed, with the remaining 99.94\% conforming exactly to the documented convention.

\subsection{Intended Use and Limitations}

MGhana-ST is intended to support research on speech recognition, speech translation, and speech representation learning for low-resource Ghanaian languages. It is not designed for forensic speaker identification or sensitive demographic inference. In addition, the short duration of many utterances limits the extent to which long-range speaker characteristics can be captured.

\section{Experimental Setup}

Our task is speech-to-English translation. We use the encoder of Whisper-small \citep{radford2023robust} (244M parameters in the full model) as the speech backbone. Input speech is converted to 80-channel log-Mel spectrograms and passed through the 12-layer Whisper-small encoder; the resulting 768-dimensional states are linearly projected to $d_{\text{model}}=256$ and consumed by a custom 4-layer Transformer decoder with 8 attention heads, trained from scratch, which autoregressively produces English text.

To adapt the encoder in the low-resource regime, we freeze its lower layers and fine-tune only the top $n$ layers, with $n$ chosen by a preliminary sweep on Ga (Appendix~\ref{sec:sweep}). The decoder is trained from random initialisation in all configurations.

The overall training objective is
\begin{equation}
\mathcal{L} =
\mathcal{L}_{\text{trans}} + \alpha\,\mathcal{L}_{\text{LID}},
\label{eq:lid_loss}
\end{equation}
\noindent
where $\alpha$ weights an auxiliary language-identification objective. The monolingual (M1) and multilingual baseline (M2) configurations use $\alpha=0$, so they optimise the translation loss alone; M4 sets $\alpha=0.1$. M3 instead applies Domain-Adversarial Training (DANN) with a gradient-reversal layer and $\lambda=0.1$; its domain-classification term is not represented by Eq.~\ref{eq:lid_loss}. Auxiliary objective variants (M3 and M4) are reported in Appendix~\ref{sec:aux-results}.

\subsection{Reproducibility and Seed Protocol}
\label{sec:reproducibility}

Two distinct random seeds govern our experiments.

The \emph{split seed} controls the audio-level assignment of recordings to train, validation, and test. It is fixed at 42 for every experiment in this paper. All runs therefore share a single, identical data partition, which is published as the manifest described in Section~\ref{sec:subset-definition}.

The \emph{training seed} controls parameter initialisation, batch ordering, and dropout. Where we report means and standard deviations, they are computed over training seeds $\{1, 2, 3\}$ with the split held fixed. The variance we report is thus training variance under a fixed partition.

Runs are not bitwise deterministic (\texttt{deterministic: false}), so exact per-run reproduction is not guaranteed even at a fixed seed; the reported means and standard deviations are the reproducible quantities.

\paragraph{Which results carry error bars.}
The M1 and M2 configurations are run at three seeds and reported as mean $\pm$ sample standard deviation. The auxiliary conditioning experiments and the cross-lingual transfer matrix are \emph{single-run} at training seed 42 and are reported as such (Appendices~\ref{sec:appendix-transfer} and~\ref{sec:aux-results}). We do not compute deltas between multi-seed and single-run configurations, as this would confound treatment effects with the particular seed draw.

\subsection{Dataset}

Main experiments use the four-language experimental subset described in Table~\ref{tab:full-data}, comprising 12.78 hours of training audio across Twi (6.21h), Ga (2.48h), Ewe (2.47h), and Fante (1.62h). All splits are partitioned at the audio-file level to prevent overlap between training and evaluation.

\paragraph{Baseline scope.}
We do not include a cascaded ASR$\rightarrow$MT baseline, because no ASR system we are aware of supports Ga, Twi, Ewe, or Fante at acceptable accuracy. A cascade is therefore not realisable as a deployable system in this setting, which is characteristic of truly low-resource languages and motivates end-to-end speech translation.

\subsection{Model Architecture}

\begin{figure}[t]
\centering
\includegraphics[width=\columnwidth]{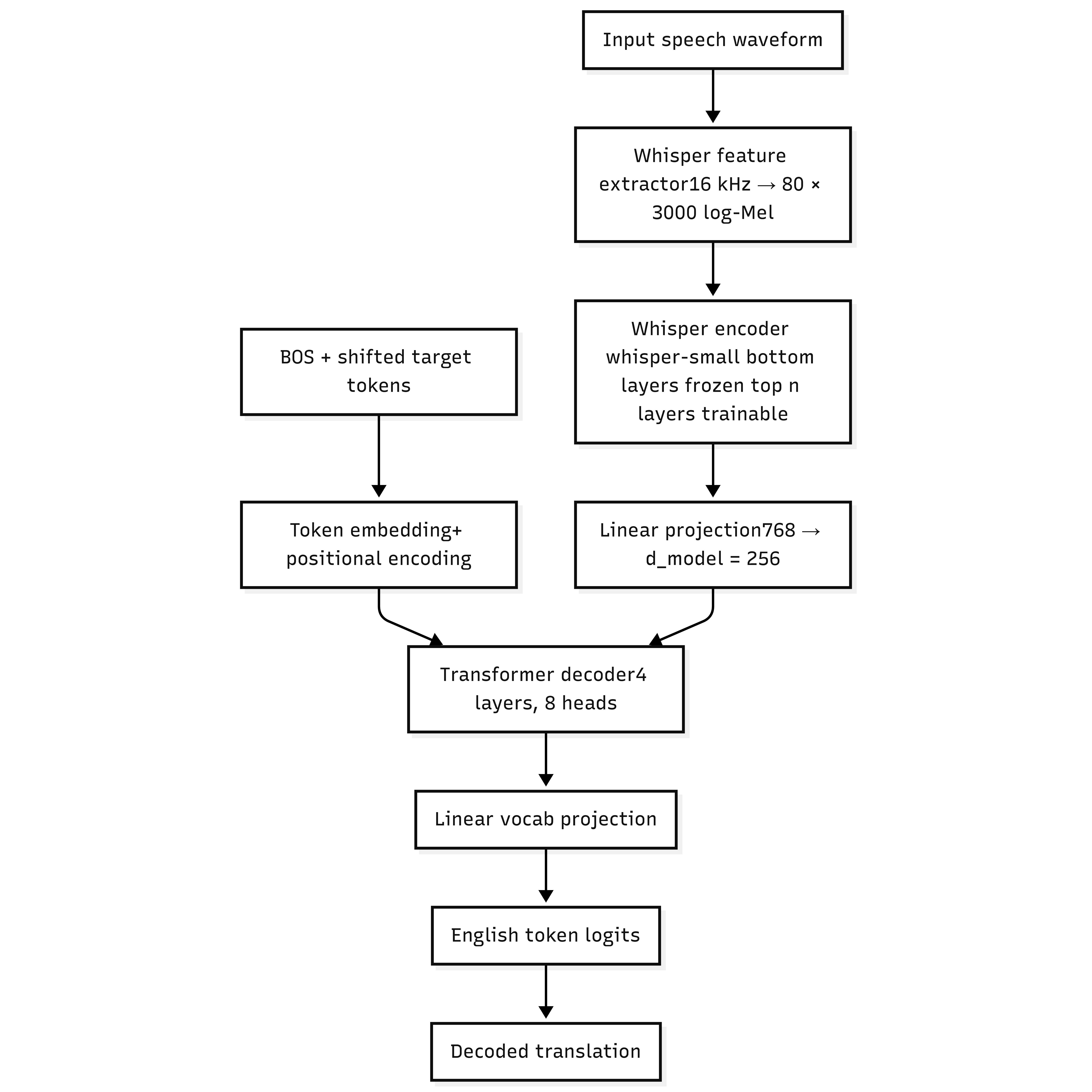}
\caption{MGhana-ST inference graph. Input speech is processed by the Whisper feature extractor and encoder, projected to the decoder dimension, and decoded autoregressively to produce an English translation.}
\label{fig:inference}
\end{figure}

Figure~\ref{fig:inference} summarises the architecture. The top $n=4$ Whisper encoder layers are fine-tuned while lower layers are frozen, selected to balance retention of pretrained acoustic structure with task-specific adaptation (see Appendix~\ref{sec:sweep} for the depth selection analysis).

\subsection{Linguistic Relatedness Ordering}
\label{sec:relatedness}

To compare different notions of relatedness within our language set, we consider URIEL syntactic similarity \citep{littell2017uriel}, geographic distance, and empirical cross-lingual transfer BLEU. Table~\ref{tab:relatedness} summarises these measures for all language pairs; the full pairwise transfer matrix is reported in Appendix~\ref{sec:appendix-transfer}.

The measures do not fully agree. URIEL ranks Ga--Twi as the closest pair, whereas transfer BLEU identifies Twi--Fante as the strongest pair. Because transfer BLEU is derived directly from model behaviour on the target task, we use it as the primary ordering signal in our analysis. The gap between Twi--Fante (18.15 BLEU) and every other pair (0.13--1.23) is more than an order of magnitude, far larger than the seed variance measured elsewhere in this paper. The divergence among these measures suggests that general typological similarity does not necessarily correspond to task-specific transfer behaviour in low-resource speech translation.

Transfer BLEU should not be interpreted as reflecting linguistic relatedness alone. It may also be influenced by factors such as dataset size, speaker variability, recording conditions, and domain overlap. Notably, all Ewe data comes from a different source corpus than the other three varieties, so domain and channel similarity is partially confounded with linguistic relatedness. We therefore treat transfer BLEU as a task-specific empirical measure that aggregates linguistic, acoustic, and domain similarity.

\begin{table}[t]
\centering
\small
\begin{tabular}{lccc}
\hline
\textbf{Pair} & \textbf{URIEL} & \textbf{Geo.} & \textbf{Transfer} \\
              & \textbf{sim. $\uparrow$} & \textbf{dist. $\downarrow$} & \textbf{BLEU $\uparrow$} \\
\hline
Twi -- Fante  & 0.772 & 0.109 & \textbf{18.15} \\
Ga -- Ewe     & 0.789 & \textbf{0.087} & 0.13 \\
Ga -- Twi     & \textbf{0.874} & 0.119 & 0.68 \\
Twi -- Ewe    & 0.794 & 0.165 & 0.14 \\
Ga -- Fante   & 0.767 & 0.211 & 1.23 \\
Ewe -- Fante  & 0.772 & 0.218 & 0.18 \\
\hline
\end{tabular}
\caption{Pairwise language relatedness across complementary measures. Transfer BLEU is the mean of the two directional zero-shot scores for each pair, single-run at seed 42. URIEL identifies Ga--Twi as closest, while transfer BLEU identifies Twi--Fante as the strongest transfer pair.}
\label{tab:relatedness}
\end{table}

\section{Results and Analysis}
\label{sec:results}

\subsection{Does Multilingual Training Help?}
\label{sec:multi-help}

Table~\ref{tab:m1-m2} reports monolingual (M1) and flat multilingual (M2) performance as means over three training seeds. The answer, at this data scale, is no: multilingual training does not reliably improve any of the four varieties, and substantially degrades two.

\paragraph{Ga and Twi are unchanged.}
Ga gains $+0.51$ BLEU and Twi $+0.06$. Both are small relative to their monolingual seed standard deviations ($1.63$ and $2.20$ respectively). We therefore treat both as null results rather than as gains.

\paragraph{Ewe and Fante degrade.}
Ewe falls from $15.48$ to $8.49$ BLEU ($-6.99$), and Fante from $47.61$ to $42.50$ ($-5.11$). Both effects exceed twice the larger of the two conditions' standard deviations and are corroborated by chrF ($-8.64$ and $-3.89$). These are the substantive findings of this section.

\paragraph{Which varieties pay.}
The two varieties harmed are the typologically isolated one and the least-resourced one. Ewe is the only Gbe variety in the set, and additionally the only variety drawn from a different source corpus, so it shares neither family structure nor domain with the majority of the training pool. Fante, at $1.62$h, has the smallest training allocation of the four, roughly a quarter of Twi's $6.21$h. Twi, the largest, is unaffected. This is the pattern expected if joint training under fixed capacity reallocates representational resources toward the better-represented and better-connected varieties.

\begin{table}[t]
\centering
\small
\setlength{\tabcolsep}{3pt}
\begin{tabular}{lccr}
\hline
\textbf{Lang.} & \textbf{M1 (mono)} & \textbf{M2 (multi)} & \textbf{$\Delta$} \\
\hline
\multicolumn{4}{l}{\textit{BLEU}} \\
Ga    & 35.25\,$\pm$\,1.63 & 35.76\,$\pm$\,0.35 & $+0.51$ \\
Twi   & 18.79\,$\pm$\,2.20 & 18.85\,$\pm$\,0.07 & $+0.06$ \\
Ewe   & 15.48\,$\pm$\,0.56 &  8.49\,$\pm$\,1.15 & $\bm{-6.99}$ \\
Fante & 47.61\,$\pm$\,0.79 & 42.50\,$\pm$\,1.43 & $\bm{-5.11}$ \\
\hline
\multicolumn{4}{l}{\textit{chrF}} \\
Ga    & 48.55\,$\pm$\,3.30 & 49.50\,$\pm$\,0.79 & $+0.95$ \\
Twi   & 30.86\,$\pm$\,0.95 & 30.58\,$\pm$\,0.84 & $-0.28$ \\
Ewe   & 34.34\,$\pm$\,0.10 & 25.70\,$\pm$\,1.86 & $\bm{-8.64}$ \\
Fante & 61.75\,$\pm$\,0.54 & 57.86\,$\pm$\,1.45 & $\bm{-3.89}$ \\
\hline
\end{tabular}
\caption{Monolingual (M1) versus flat multilingual (M2) performance per variety, mean $\pm$ sample standard deviation over three training seeds with the data split held fixed. Bold marks deltas whose magnitude exceeds twice the larger of the two standard deviations. Corpus-level M2 BLEU is $22.50 \pm 0.27$. Per-seed values are in Appendix~\ref{sec:appendix-seeds}.}
\label{tab:m1-m2}
\end{table}

\subsection{Seed Variance and the Cost of Single-Run Reporting}
\label{sec:seed-variance}

The seed standard deviations in Table~\ref{tab:m1-m2} are themselves informative, and they differ between configurations in a way that matters for how such comparisons are reported.

For Ga and Twi, the monolingual models are markedly less stable than the multilingual ones: $\pm1.63$ versus $\pm0.35$ BLEU for Ga, and $\pm2.20$ versus $\pm0.07$ for Twi. For Ewe and Fante the ordering reverses ($\pm0.56$ versus $\pm1.15$, and $\pm0.79$ versus $\pm1.43$). The multilingual model sees two to eight times as much training audio per epoch as any single monolingual model, which plausibly stabilises Ga and Twi. That the two degraded varieties are also the ones whose multilingual runs vary most is consistent with their performance depending on how shared capacity is allocated in each run, although we have not tested this directly.

This matters methodologically. Where the monolingual condition is the noisier one, a single-run M1--M2 comparison pairs a noisy monolingual number with a stable multilingual one, so the sign of the reported delta is largely determined by where the monolingual draw happens to fall. An earlier single-run version of this analysis, at training seed 42, reported $+2.95$ (Ga), $+1.68$ (Twi), $-4.46$ (Ewe), and $+0.46$ (Fante), that is, three gains and one loss. Under replication, all three apparent gains disappear, and Fante's reverses sign.

We do not believe this failure mode is specific to our setup. Low-resource multilingual comparisons are often reported from a single run, and wherever monolingual baselines trained on a few hours of audio are less stable than the pooled model, as they are for Ga and Twi here, single-run reporting can produce apparent positive transfer from seed variation alone. Our results suggest that comparisons at this data scale should report seed variance for both conditions as a matter of course.

\subsection{Separating Family from Domain}

Because all Ewe data is drawn from UGSpeechData while Ga, Twi, and Fante come entirely from the Financial Inclusion Speech Dataset, the Ewe degradation admits two explanations: family distance (Gbe vs.\ Akan) and domain mismatch. Ga provides a partial control, since it shares corpus provenance with Twi and Fante but does not share their language family. If domain similarity alone drove transfer, Ga should transfer strongly to Twi and Fante; instead Ga$\rightarrow$Twi and Ga$\rightarrow$Fante reach only 0.98 and 1.48 BLEU (Table~\ref{tab:transfer-full}, Appendix~\ref{sec:appendix-transfer}), far below Twi$\rightarrow$Fante and Fante$\rightarrow$Twi (23.36 and 12.94). Domain overlap alone is therefore not sufficient to generate transfer in this dataset.

\subsection{Effect of Linguistic Relatedness}

To examine the role of linguistic relatedness, we compare multilingual outcomes with pairwise relatedness measures and cross-lingual transfer scores. The strongest zero-shot transfer is observed between Twi and Fante, which is consistent with their closer linguistic relationship and shared corpus provenance. By contrast, language pairs involving Ewe and Ga show substantially weaker transfer.

The multi-seed results qualify how this should be read. Strong pairwise transfer between Twi and Fante does not translate into Fante benefiting from joint training; Fante is one of the two varieties that degrades. High measured transfer therefore predicts that a language \emph{can} draw on a related pool, not that it will be allocated capacity within it. Transfer BLEU remains the more direct task-specific indicator of which pairs interact in this setting, but it is not by itself a predictor of which languages gain under joint training.

These comparisons suggest that, in our setting, task-specific transfer measurements are more informative than general typological similarity for anticipating which languages interact.

\subsection{Target-Side Overlap and Reference Diversity}
\label{sec:overlap}

Because MGhana-ST is drawn from narrow-domain source corpora, English reference translations repeat across examples, which inflates BLEU without reflecting translation capability. \textbf{Cross-language score comparisons are not like-for-like.} Fante's high monolingual score ($47.61$ BLEU from 1.62h of training audio) is likely influenced by reference repetition rather than indicating that Fante is an easier language: only 323 distinct English strings cover its 524 test utterances. Ewe sits at the opposite extreme, with 1.5\% target overlap and 3.31\,s mean duration, so its lower absolute BLEU ($15.48$) reflects a harder generation task. Absolute BLEU should therefore not be compared across the four varieties. This does not affect the within-language comparisons that carry our main findings: the M1--M2 contrast holds the test set fixed for each language, so the deltas are measured against each variety's own baseline.

Further details on prompt reuse and domain overlap appear in Appendix~\ref{sec:overlap-full}.

\section{Discussion}

Our results support four broader points about low-resource multilingual speech translation.

First, multilingual training should not be assumed to provide benefits at all at this data scale. In MGhana-ST, joint training left two varieties unchanged and substantially degraded the other two. Practitioners building for these varieties at comparable data volumes should treat monolingual training as the default and multilingual training as a hypothesis requiring per-language validation.

Second, the varieties that degrade share identifiable characteristics. The typologically isolated variety and the least-resourced variety degraded, while Twi, the best-resourced variety, did not. Notably, Fante degraded despite belonging to the strongest transfer pair in the set, indicating that relatedness to the training pool does not protect a language whose share of that pool is small.

Third, typology-based similarity and task-specific transfer behaviour are not interchangeable. Empirical cross-lingual transfer identifies which language pairs interact more accurately than URIEL similarity does, yet it does not predict which languages benefit under joint training.

Fourth, and methodologically, the reliability of results in this regime is itself at stake. Our own earlier single-run analysis reported the opposite headline conclusion. For the two varieties on which that analysis showed its largest gains, Ga and Twi, the monolingual models were roughly five to thirty times less stable across seeds than the multilingual model. Our results suggest that multi-seed reporting for both conditions should be considered standard practice for sub-twenty-hour multilingual comparisons.

These observations arise from experiments on MGhana-ST using Whisper-small under the training configurations considered here, and should be interpreted as empirical findings within this experimental setting rather than universal properties of multilingual speech translation.

\section*{Limitations}

\paragraph{Scale and language imbalance.}
The experimental subset contains only about 16.1 hours of paired speech and translation data. Twi accounts for the largest share (6.21h of 12.78h of training audio), while Fante is the least represented at 1.62 hours. Model comparisons are therefore sensitive to data composition.

\paragraph{Unavailable audio and subset composition.}
As reported in Section~\ref{sec:subset-definition}, 12.6\% of annotated segments reference audio not present in our working copy (39.9\% for Ewe and 30.4\% for Fante). We have not established whether these files are absent upstream or only locally, nor whether their absence correlates with speaker, recording session, or collection batch. This does not compromise the within-language M1--M2 contrasts, since all configurations train and evaluate on identical data, but it does limit what the Ewe and Fante results can be said to represent about those varieties in general.

\paragraph{Incomplete multi-seed coverage.}
The auxiliary conditioning experiments and the transfer matrix are single-run. Given seed standard deviations of up to 2.20 BLEU, these should be treated as exploratory. Extending multi-seed coverage to all configurations is a priority for future work.

\paragraph{Split variance not measured.}
All runs share a single data partition. Our error bars therefore capture training variance only. Variance across data partitions is not measured.

\paragraph{Corpus provenance.}
MGhana-ST aggregates two source corpora that differ in domain, recording channel, speaking style, and utterance duration, unevenly distributed across varieties. Language identity is therefore partially confounded with acoustic and domain characteristics. We use Ga as a partial control to bound this effect, but a clean separation would require matched-domain data for all four varieties.

\paragraph{Representational coverage.}
MGhana-ST covers four varieties from one region of Ghana and should not be taken to represent Ghanaian languages broadly. Domain coverage is narrow, dominated by financial-technology scenarios for three varieties. Dialectal coverage within Twi is not characterised (Akuapem and Asante are merged). Speaker demographic distributions for the subset are not reported.

\paragraph{Encoder depth selection.}
The encoder fine-tuning depth was chosen from a single-run sweep scored on the Ga test split of a preliminary partition (Appendix~\ref{sec:sweep}). The selected depth is held fixed across all configurations, so it does not favour M1 or M2, but it is not a held-out choice.

\paragraph{Utterance length and reference diversity.}
Three of the four varieties consist largely of single words and short phrases (mean duration 1.33--1.90\,s), and their English references repeat substantially. MGhana-ST does not support conclusions about sentence-level speech translation, and absolute BLEU is not comparable across its four varieties.

\paragraph{Bitwise reproducibility.}
Training runs are not bitwise deterministic, so individual runs cannot be reproduced exactly even at a fixed seed. The reported means and standard deviations are the reproducible quantities.

\section*{Use of AI Assistants}

AI assistants were used during the preparation of this paper for coding support, editorial refinement, and language editing. They were not used to generate experimental results, analyses, or conclusions, all of which are the authors' own.

\section*{Acknowledgments}

\ifanonymous
  [Withheld for review.]
\else
  We thank the annotators and collaborators who contributed to data collection, translation, and dataset preparation. We also acknowledge support from Ashesi University and AdwumaTech AI.
\fi

\bibliography{custom}

\appendix
\section*{Appendix}

\section{Encoder Fine-tuning Depth Selection}
\label{sec:sweep}

Before the main experiments, we performed a sweep over the number of unfrozen top Whisper-small encoder layers, with $n \in \{0, 2, 4, 6, 12\}$, on the Ga subset only, using a preliminary partition (2.46h train, 0.29h val, 0.30h test) that predates the final experimental subset. Scores are on that partition's test split.

Performance is relatively stable across $n \in \{0,2,4,6\}$, with BLEU highest under full freezing ($n=0$, 36.75) and chrF highest at $n=4$ (53.31). Unfreezing all 12 layers substantially reduces both metrics (BLEU 29.04, chrF 44.53), consistent with degradation of pretrained representations under extensive fine-tuning in a low-resource setting. We select $n=4$ as a practical operating point because it achieves the highest chrF while maintaining BLEU close to the strongest observed values. Two caveats apply: the sweep predated preparation of the full multilingual data, and it is single-run, so the BLEU spread across $n \in \{0,2,4,6\}$ (1.54) is smaller than the monolingual Ga seed standard deviation (1.63). The ranking among moderate depths is not resolvable from this sweep.

\begin{table}[ht]
\centering
\begin{tabular}{lcc}
\hline
\textbf{Unfrozen layers} ($n$) & \textbf{BLEU} & \textbf{chrF} \\
\hline
0  & 36.75 & 49.59 \\
2  & 35.21 & 50.81 \\
4$^{\ast}$ & 36.25 & 53.31 \\
6  & 36.63 & 49.45 \\
12 & 29.04 & 44.53 \\
\hline
\end{tabular}
\caption{Encoder fine-tuning depth sweep on the Ga test split of the preliminary partition (single-run). $^{\ast}$Depth selected for main experiments.}
\label{tab:sweep}
\end{table}

\section{Segment Retention Audit}
\label{sec:exclusion}

Not every annotated segment enters the experimental subset (Table~\ref{tab:retention}). The largest exclusion is event-only segments, whose translation consists solely of a bracketed event tag with no accompanying words: 48{,}811 segments (53.9\% of the 90{,}583 annotated segments), at rates of 66.2\% for Ga, 59.7\% for Twi, 46.0\% for Fante, and 29.5\% for Ewe. The second is unavailable audio (11{,}445 segments, 12.6\%), concentrated in Ewe (39.9\%) and Fante (30.4\%). Thirteen further segments are excluded for other reasons, leaving 30{,}314 retained segments. The full annotation set also contains segments labelled with languages other than the four varieties studied here; these are outside the experimental subset and are not counted in Table~\ref{tab:retention}.

\begin{table}[ht]
\centering
\small
\setlength{\tabcolsep}{3.5pt}
\begin{tabular}{lrrrr}
\hline
\textbf{Category} & \textbf{Ga} & \textbf{Twi} & \textbf{Ewe} & \textbf{Fante} \\
\hline
Annotated segs.   & 19{,}370 & 37{,}100 & 11{,}333 & 22{,}780 \\
Audio missing     & 0 & 12 & 4{,}516 & 6{,}917 \\
Event-only        & 12{,}822 & 22{,}156 & 3{,}346 & 10{,}487 \\
Other             & 7 & 3 & 3 & 0 \\
\hline
\textbf{Retained} & \textbf{6{,}541} & \textbf{14{,}929} & \textbf{3{,}468} & \textbf{5{,}376} \\
Retention rate    & 33.8\% & 40.2\% & 30.6\% & 23.6\% \\
\hline
\end{tabular}
\caption{Segment retention by exclusion category. Ten retained segments (Twi 3, Ewe 5, Fante 2) do not appear in the final splits, so the totals in Table~\ref{tab:corpus-stats} are slightly lower.}
\label{tab:retention}
\end{table}

\section{Cross-Lingual Transfer BLEU Matrix}
\label{sec:appendix-transfer}

Table~\ref{tab:transfer-full} reports the full cross-lingual transfer evaluation, in which each M1 monolingual model is evaluated on all four language test sets. These are single-run results at seed 42.

\begin{table}[ht]
\centering
\resizebox{\columnwidth}{!}{%
\begin{tabular}{lcccccccc}
\hline
\multirow{2}{*}{\textbf{Model}}
  & \multicolumn{2}{c}{\textbf{Ga}}
  & \multicolumn{2}{c}{\textbf{Twi}}
  & \multicolumn{2}{c}{\textbf{Ewe}}
  & \multicolumn{2}{c}{\textbf{Fante}} \\
\cmidrule(lr){2-3}\cmidrule(lr){4-5}\cmidrule(lr){6-7}\cmidrule(lr){8-9}
& BLEU & chrF & BLEU & chrF & BLEU & chrF & BLEU & chrF \\
\hline
M1-ga    & \textbf{34.76} & \textbf{49.25} & 0.98  & 8.25  & 0.07  & 7.13  & 1.48  & 9.24  \\
M1-twi   & 0.38  & 8.11  & \textbf{17.91} & \textbf{29.70} & 0.17  & 9.21  & 23.36 & 43.17 \\
M1-ewe   & 0.18  & 8.06  & 0.11  & 9.87  & \textbf{14.06} & \textbf{31.01} & 0.23  & 6.94  \\
M1-fante & 0.97  & 9.06  & 12.94 & 23.72 & 0.13  & 9.74  & \textbf{45.73} & \textbf{59.83} \\
\hline
\end{tabular}%
}
\caption{Cross-lingual transfer matrix: each M1 model is evaluated on all four language test sets (single-run, seed 42). Bold diagonal entries are in-language results. Off-diagonal entries are zero-shot transfers.}
\label{tab:transfer-full}
\end{table}

\section{Auxiliary Conditioning}
\label{sec:aux-results}

These results are exploratory and single-run at seed 42. Because Section~\ref{sec:seed-variance} establishes that single-run differences at this data scale can exceed the effects of interest, we report them as directional evidence only. All comparisons are made against the single-run M2 baseline at seed 42 (Ga 37.71, Twi 19.59, Ewe 9.60, Fante 46.19; corpus 23.72).

We evaluate two auxiliary objectives. M3 applies Domain-Adversarial Training (DANN) with gradient reversal to encourage language-invariant encoder representations ($\lambda=0.1$). M4 applies Language Identification (LID) supervision to encourage language-discriminative representations ($\alpha=0.1$ in Equation~\ref{eq:lid_loss}).

\paragraph{Invariance redistributes.}
DANN improves Ewe substantially, from 9.60 to 13.23 BLEU ($+3.63$). However, Fante declines by 4.57 BLEU and Twi by 0.83, while Ga is unchanged. Corpus-level BLEU is slightly below M2, so DANN reallocates performance rather than improving it uniformly.

\paragraph{Discriminativeness does not help.}
LID conditioning underperforms M2 for all four languages (Ga $-1.59$, Fante $-1.76$, Ewe $-0.64$, Twi $-0.51$), with corpus BLEU 0.76 below M2. Encouraging the encoder to represent language identity more sharply does not mitigate interference at this seed.
We have not verified that either objective changes what the encoder represents; probing the encoder states for language identity, in the manner of prior probing work on distributed representations \citep{acquaye2023hypernym}, would test this directly.

\begin{table}[ht]
\centering
\small
\setlength{\tabcolsep}{3pt}
\begin{tabular}{lccccc}
\hline
\textbf{Run} & \textbf{Ga} & \textbf{Twi} & \textbf{Ewe} & \textbf{Fante} & \textbf{Corpus} \\
\hline
\multicolumn{6}{l}{\textit{BLEU}} \\
M2 (multi)  & 37.71 & 19.59 & 9.60  & 46.19 & 23.72 \\
M3 (DANN)   & 37.74 & 18.76 & 13.23 & 41.62 & 23.38 \\
M4 (LID)    & 36.12 & 19.08 & 8.96  & 44.43 & 22.96 \\
\hline
\end{tabular}
\caption{Auxiliary conditioning results (single-run, seed 42). All comparisons are against M2 at the same seed, not against the three-seed mean in the main text.}
\label{tab:m3-m4}
\end{table}

\section{Target-Side Overlap and Prompt Reuse}
\label{sec:overlap-full}

Reference repetition has two consequences. First, Fante reference repetition is a direct consequence of the collection protocol: speakers of a given dialect recorded a shared list of approximately 130 prompt sentences. Because our splits are constructed at the audio-file level, a prompt sentence recorded by one speaker may appear in training while the same sentence recorded by another appears in test. Scores on Ga, Twi, and Fante reflect a setting in which the space of target sentences is small and partially shared across splits, and should be read as performance on a closed prompt inventory. Ewe, drawn from image-description recordings, is not subject to this and shows correspondingly low reference repetition (1.5\%).

Second, reference repetition raises a natural objection to using transfer BLEU as a relatedness signal (Section~\ref{sec:relatedness}): if Twi and Fante share a domain and a stock of high-frequency English formulae, strong Twi--Fante transfer might reflect a shared phrase inventory. Ga shares the same source corpus and domain as Twi and Fante, with comparable reference repetition (27.2\% for Ga against 27.4\% for Twi), yet Ga--Twi transfer is negligible (0.98 and 0.38 BLEU). Domain overlap and reference repetition are thus not sufficient to generate transfer; the Akan relationship shared by Twi and Fante is what distinguishes the two cases.

\section{Per-Seed Results}
\label{sec:appendix-seeds}

Table~\ref{tab:per-seed} reports the individual seed results underlying the means in Table~\ref{tab:m1-m2}.

\begin{table}[ht]
\centering
\small
\setlength{\tabcolsep}{4pt}
\begin{tabular}{llcccc}
\hline
\textbf{Cfg} & \textbf{Seed} & \textbf{Ga} & \textbf{Twi} & \textbf{Ewe} & \textbf{Fante} \\
\hline
\multicolumn{6}{l}{\textit{BLEU}} \\
M1 & 1 & 36.83 & 17.25 & 16.10 & 46.82 \\
M1 & 2 & 33.57 & 21.31 & 15.31 & 47.60 \\
M1 & 3 & 35.35 & 17.80 & 15.02 & 48.40 \\
M2 & 1 & 35.98 & 18.78 &  7.21 & 41.48 \\
M2 & 2 & 35.35 & 18.92 &  9.44 & 44.13 \\
M2 & 3 & 35.94 & 18.85 &  8.81 & 41.89 \\
\hline
\end{tabular}
\caption{Per-seed BLEU for M1 and M2, split held fixed. Corpus-level M2 BLEU is 22.30, 22.81, and 22.39 for seeds 1--3.}
\label{tab:per-seed}
\end{table}

\section{Training Details}

Training hyperparameters: batch size 16 for 60 epochs, learning rate 0.0005, 500 warmup steps, gradient clipping 1.0. Checkpoints saved every 5 epochs. Top 4 Whisper-small encoder layers fine-tuned; lower layers frozen. Decoder trained from scratch. Explicit language conditioning disabled for M1/M2 baselines. The full 12-layer Whisper-small encoder is used as the speech backbone. Decoding uses beam size 4. The annotation state corresponds to commit \texttt{4996b709abd584464a6c\allowbreak fb7bd256504fd5eed7a0} (25 March 2026). Environment: Python 3.12.12, PyTorch 2.10.0+cu126, Transformers 5.1.0, on a single NVIDIA A100-SXM4-40GB GPU.

\end{document}